\documentclass{article}

\usepackage[preprint]{corl_2025} %

\usepackage{array}
\usepackage{amssymb}
\usepackage{amsmath}
\usepackage{amsfonts}
\usepackage{booktabs}
\usepackage{colortbl}
\usepackage{enumitem}
\usepackage{graphicx}
\usepackage{verbatim}
\usepackage{wrapfig}
\usepackage{xspace}
\usepackage{multirow}
\usepackage[scaled=0.85]{beramono}
\usepackage[T1]{fontenc}
\newcolumntype{P}[1]{>{\centering\arraybackslash}p{#1}}
\usepackage{caption}
\usepackage{siunitx} 
\usepackage{color, soul} 

\newcommand{\method}{D-CODA\xspace}
\newcommand{\ba}{\mathbf{a}}

\definecolor{lightgreen}{rgb}{0.8, 1.0, 0.8}

\definecolor{figmaroon}{rgb}{0.776, 0.184, 0.353}
\definecolor{figgreen}{rgb}{0.255, 0.506, 0.498}
\definecolor{figyellow}{rgb}{0.961, 0.749, 0.353}
\definecolor{figblue}{rgb}{0.592, 0.796, 0.843}
\definecolor{figred}{rgb}{0.933, 0.271, 0.251}

\title{\method: Diffusion for Coordinated\\Dual-Arm Data Augmentation}

\author{
  I-Chun Arthur Liu \quad
  Jason Chen \quad
  Gaurav S. Sukhatme\thanks{GSS holds concurrent appointments as a Professor at USC and as an Amazon Scholar. This paper describes work performed at USC and is not associated with Amazon.} \quad Daniel Seita  \\
  Department of Computer Science, University of Southern California
}

\begin{document}
\maketitle

\vspace{-15pt}
\begin{abstract}
Learning bimanual manipulation is challenging due to its high dimensionality and tight coordination required between two arms. Eye-in-hand imitation learning, which uses wrist-mounted cameras, simplifies perception by focusing on task-relevant views. However, collecting diverse demonstrations remains costly, motivating the need for scalable data augmentation. While prior work has explored visual augmentation in single-arm settings, extending these approaches to bimanual manipulation requires generating viewpoint-consistent observations across both arms and producing corresponding action labels that are both valid and feasible. In this work, we propose Diffusion for COordinated Dual-arm Data Augmentation (\method), a method for offline data augmentation tailored to eye-in-hand bimanual imitation learning that trains a diffusion model to synthesize novel, viewpoint-consistent wrist-camera images for both arms while simultaneously generating joint-space action labels. It employs constrained optimization to ensure that augmented states involving gripper-to-object contacts adhere to constraints suitable for bimanual coordination. We evaluate \method on 5 simulated and 3 real-world tasks. Our results across 2250 simulation trials and 300 real-world trials demonstrate that it outperforms baselines and ablations, showing its potential for scalable data augmentation in eye-in-hand bimanual manipulation.
Our project website is at: \href{https://dcodaaug.github.io/D-CODA/}{https://dcodaaug.github.io/D-CODA/}.
\end{abstract}


\section{Introduction}

Bimanual robotic manipulation is often necessary for diverse real-world tasks~\cite{Bimanual_Taxonomy_2022}.
Recently, researchers have shown the merits of wrist cameras in visual-based robot learning for manipulation~\cite{hsu2022visionbasedmanipulatorsneedhands,kim2023givingrobotshandlearning,young2020visual}, as they help simplify certain aspects of the visual scene and focus on task-relevant objects.
However, a fundamental challenge remains: learning-based systems still require large amounts of data for effective generalization, and collecting additional data across different viewpoints and states is both costly and labor-intensive.

One way to address this issue is with data augmentation. This is a widely used technique in computer vision~\cite{simard2003best,krizhevsky2012alexnet} and visual reinforcement learning~\cite{laskin2020reinforcementlearningaugmenteddata,kostrikov2021imageaugmentation} to broaden the training data and facilitate generalization.
In robotics, prior work has explored ways to automatically generate and synthesize novel image views while preserving action labels~\cite{zhang2024diffusionmeetsdagger,zhou2023nerfpalmhand,chen2024roviaug,tian2024vista}, although such efforts have been limited to single-arm settings.
Bimanual manipulation introduces additional challenges, including higher degrees of freedom (DOFs), enforcing consistency across the two generated wrist-camera views, and ensuring that augmented actions remain valid for coordinated manipulation.
A complementary approach to increase data coverage is Dataset Aggregation (DAgger)~\cite{ross2011dagger}, which leverages a supervisor to provide corrective labels. However, this method incurs additional online environment interactions and assumes a supervisor is available, which is not always feasible.

\begin{figure*}[t]
\center
\includegraphics[width=1.0\textwidth]{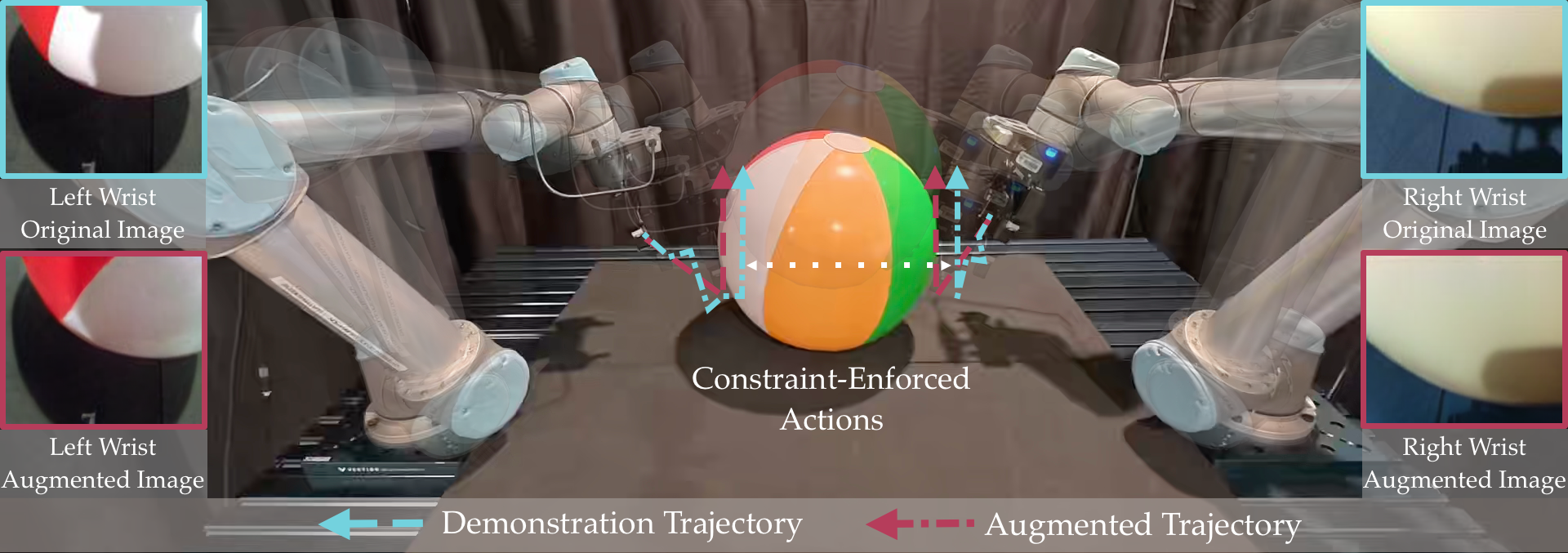} 
\caption{
Overview of \method for a coordinated bimanual lifting task with two UR5 arms. \method is a method for offline data augmentation in bimanual eye-in-hand imitation learning. Given a pair of wrist camera images from demonstrations and sampled pose perturbations, a diffusion model generates novel and viewpoint-consistent wrist images for both arms (i.e., consistent object shape, color, position, and orientation). We use an optimization procedure to generate constraint-enforced actions to ensure augmented states are appropriate for bimanual coordination. This enables scalable data augmentation of diverse training data. 
}
\label{fig:pull}
\vspace*{-10pt}
\end{figure*}

In this paper, we propose 
\textbf{D}iffusion for \textbf{CO}ordinated \textbf{D}ual-arm Data \textbf{A}ugmentation (\textbf{\method}), a diffusion-based data augmentation framework tailored for eye-in-hand bimanual imitation learning. \method synthesizes novel wrist-camera views along with consistent action labels to generate additional training data for bimanual manipulation policies \emph{offline}, without the need for a simulator or the recreation of experimental setups. We design a diffusion model that takes two reference wrist images and camera pose perturbations as input and synthesizes novel viewpoint-consistent wrist-camera images for both arms. We leverage a Large Vision Model, SAM2~\cite{ravi2024sam}, to decompose any bimanual manipulation task into contactless (free-space) and contact-rich states. For contactless states, we uniformly sample random camera pose perturbations, while for contact-rich states, we employ constrained optimization to ensure that the perturbations satisfy coordination constraints required for bimanual manipulation. See Figure~\ref{fig:pull} for an overview. 

Our contributions are as follows:
(i) A novel method for bimanual manipulation that leverages diffusion models to generate diverse and consistent wrist camera images. 
(ii) A perception-based pipeline that decomposes any bimanual manipulation task into contactless and contact-rich states. For contact-rich states, we introduce a camera perturbation sampling procedure that generates constraint-enforced action labels.
(iii) Experiments in 5 simulation and 3 real-world tasks that demonstrate the effectiveness of \method over alternative baseline methods and ablations.

\section{Related Work}
\label{sec:related_work}

\textbf{Bimanual Manipulation.}
Bimanual manipulation~\cite{Bimanual_Taxonomy_2022} is essential for a wide range of real-world tasks that are difficult to perform with one arm, such as folding fabrics~\cite{avigal2022speedfolding,maitin2010cloth,canberk2022clothfunnels,colome2018dimensionality,mail2023,weng2021fabricflownet}, inserting objects into deformable bags~\cite{autobag2023,slipbagging2023,bagallyouneed2023}, and handling food~\cite{Grannen2022LearningBS,grannen2023stabilize}. These tasks often require tight coordination between the arms, either through simultaneous motions or an acting-stabilizing division of roles~\cite{grannen2023stabilize,liu2024voxactb} where one arm stabilizes parts of an item (e.g., holding food) to enable the other arm to act (e.g., cutting the food). 
While we mainly test our method for bimanual tasks where arms move simultaneously, our approach is not task-specific. 

Some prior work on bimanual manipulation formalizes coordination via learned primitives~\cite{batinica2017compliant} or constraint-based representations~\cite{ureche2018constraints}, but may suffer from generalization in unseen test-time scenarios. 
More general learning-based methods have emerged to address these limitations. Some rely on deep reinforcement learning (RL)~\cite{Chitnis2020,Biman_AC_2023}, which can be useful for simulation-based training of policies to control high-DOF hands~\cite{BiDexHands2022,lin2024learning,twist_lids_biman_2024,robopianist2023} or humanoids~\cite{sferrazza2024humanoidbench}. 
However, deep RL alone is generally difficult and brittle for bimanual manipulation~\cite{chernyadev2024bigym}; therefore, researchers have explored imitation learning~\cite{franzese2023interactive,xie2020,bahety2024screwmimic,shi2023waypointbased,zhou2024learningdiversebimanual,zhou2025teachoncelearnoneshot,lu2024anybimanual}. 
In a landmark paper,~\citet{Zhao-RSS-23} showed the benefit of predicting sequences of actions to learn fine-grained bimanual manipulation from demonstrations. Data scaling~\cite{liu2024rdt,black2024pi0visionlanguageactionflowmodel,pertsch2025fastefficientactiontokenization} and improved robot hardware~\cite{aloha2_2024,fu2024mobile,zhao2024alohaunleashed,wang2024dexcap,ding2024bunnyvisionpro} have enabled great improvement and generalization in bimanual manipulation. 
Despite such progress, significant room remains to achieve human-level generalization, and methods still struggle when facing novel viewpoints or out-of-distribution states~\cite{tian2024vista}.
Our focus is complementary and is a general data augmentation approach compatible with diverse eye-in-hand imitation systems.

\textbf{Data Augmentation in Robotics.}
Data augmentation is a widely used strategy to improve generalization in supervised learning systems such as behavioral cloning~\cite{Pomerleau_behavior_cloning}. These methods suffer from compounding execution errors at test time, where small prediction errors lead to out-of-distribution states that result in larger errors~\cite{ross2011dagger}. 
Data augmentation techniques in robotics can be roughly divided into \emph{environment-level} augmentation and \emph{trajectory-level} augmentation. 
Environment-level methods aim to expand visual diversity or semantic richness of training data. These include automatic environment generation using LLMs~\cite{wang2024gen,huagensim2,katara2024gen2sim,wang2024robogen} and controllable visual and scene augmentation~\cite{chen2024semanticallycontrollable,yuan2025roboengineplugandplayrobotdata,bharadhwaj2024roboagent}. 
Some works also synthesize image-keypoint pairs~\cite{tangandrajkumar2025kalie} or hand-object interactions~\cite{ye2023affordancediffusion}. These techniques are complementary, as we study imitation learning from existing offline RGB trajectory data.
More closely related works include RoVi-Aug~\cite{chen2024roviaug} and VISTA~\cite{tian2024vista}, which use diffusion models to generate novel viewpoints but lack action label supervision. In contrast, \method generates viewpoint-consistent images and corresponding joint-space action labels.

Trajectory-level augmentation methods~\cite{Laskey2017DARTNI,mandlekar2023mimicgen,garrett2024skillmimicgen,jiang2025dexmimicen} synthesize new robot states, transitions, and/or actions. 
MimicGen~\cite{mandlekar2023mimicgen}, SkillMimicGen~\cite{garrett2024skillmimicgen}, and DexMimicGen~\cite{jiang2025dexmimicen} generate full demonstration trajectories, but rely on access to simulation or environment interaction during data generation, whereas \method operates offline. 
Other works augment state-based inputs~\cite{mitrano2022dataaugmentationmanipulation,ke2024ccil} which limits applicability to vision-based learning.
\citet{zhou2023nerfpalmhand} use NeRF~\cite{mildenhall2020nerf} to augment visual input for corrective imitation but assume static scenes. 
Among the most closely related approaches is Diffusion Meets DAgger (DMD)~\cite{zhang2024diffusionmeetsdagger}, which augments single-arm eye-in-hand images with action labels using a diffusion model~\cite{Yu2023PhotoconsistentNVS}. \method builds on this foundation by demonstrating how to extend it to bimanual setups through a unified framework that synthesizes left and right wrist-camera views and employs constrained optimization to generate action labels suitable for bimanual manipulation.

\section{Problem Statement and Preliminaries}
\label{sec:problem}
\vspace{-6pt}

We assume a bimanual robot with a left arm $l$ and right arm $r$.
Throughout the following sections, mathematical notations with superscripts $l$ and $r$ denote the left and right arms, respectively.
We study vision-based eye-in-hand imitation learning, which trains a policy $\pi_\theta$ parameterized by $\theta$ that learns from demonstration data with wrist camera images.
To indicate the source arm for each wrist camera image, we use the $I^l$ and $I^r$ notation, though we may suppress the superscripts if the distinction is not necessary.
To represent images at time $t$ in a demonstration, we use $I_t^l$ and $I_t^r$. 
All images are in $\mathbb{R}^{H\times W\times 3}$ with matching height $H$ and width $W$ values. 
These form the policy input, which produces actions ${\ba_t = \pi_\theta((I_t^l, I_t^r))}$. Here, ${\ba_t = (\ba_t^l, \ba_t^r)}$, where $\ba_t^l$ and $\ba_t^r$
are target joint positions for the respective arms.
To train $\pi_\theta$, imitation learning uses a dataset of expert demonstrations ${\mathcal{D} = \{\tau_1, \ldots, \tau_M\}}$.
Each $\tau_i$ is a sequence of wrist-camera images observations and actions: ${\tau_i = (I_1^l, I_1^r, \ba_{1}^l, \ba_{1}^r, \ldots, I_T^l, I_T^r, \ba_{T}^l, \ba_{T}^r)}$ for a demonstration with $T$ time steps. 

\textbf{Synthesizing Novel Bimanual Images and Actions}: 
Our method synthesizes novel eye-in-hand viewpoint images while automatically deriving suitable actions to make the robot return to in-distribution data. Based on~\cite{zhang2024diffusionmeetsdagger}, we formalize this problem as learning a function $f_\psi$ that creates an eye-in-hand image conditioned on a current image and a pose perturbation $\Delta p$.  
In this case, let $\Delta p = {}_aT_b$ represent the pose transformation between two cameras $a$ and $b$, where $a$ is the source and $b$ is the target. To represent images from these cameras for both arms, we suppress $t$ and instead use the following notation: $\{I_a^l, I_a^r, I_b^l, I_b^r\}$. 
However, if notation requires specifying a camera $\{a,b\}$ as well as time $t$, both camera and time are included in the subscript (e.g., $I_{b,t}^l$), with the camera listed first, then the timestep.
Given the source images $I_a^l$ and $I_a^r$ and pose transformations $\Delta p^{l}$ and $\Delta p^{r}$ as input, $f_\psi$ must synthesize novel and consistent images $\tilde{I}_b^l$ and $\tilde{I}_b^r$ for the two cameras, matching the targets $I_b^l$ and $I_b^r$. Additionally, we use $\Delta p$ to compute perturbed actions ${\tilde{\ba}_t = (\tilde{\ba}_t^l, \tilde{\ba}_t^r) }$. Finally, an augmented dataset of novel viewpoints with corresponding action labels, $\tilde{\mathcal{D}}$, is generated.

\section{Method: \method}
\label{sec:method}

\begin{figure*}[t]
\center
\includegraphics[width=1.0\textwidth]{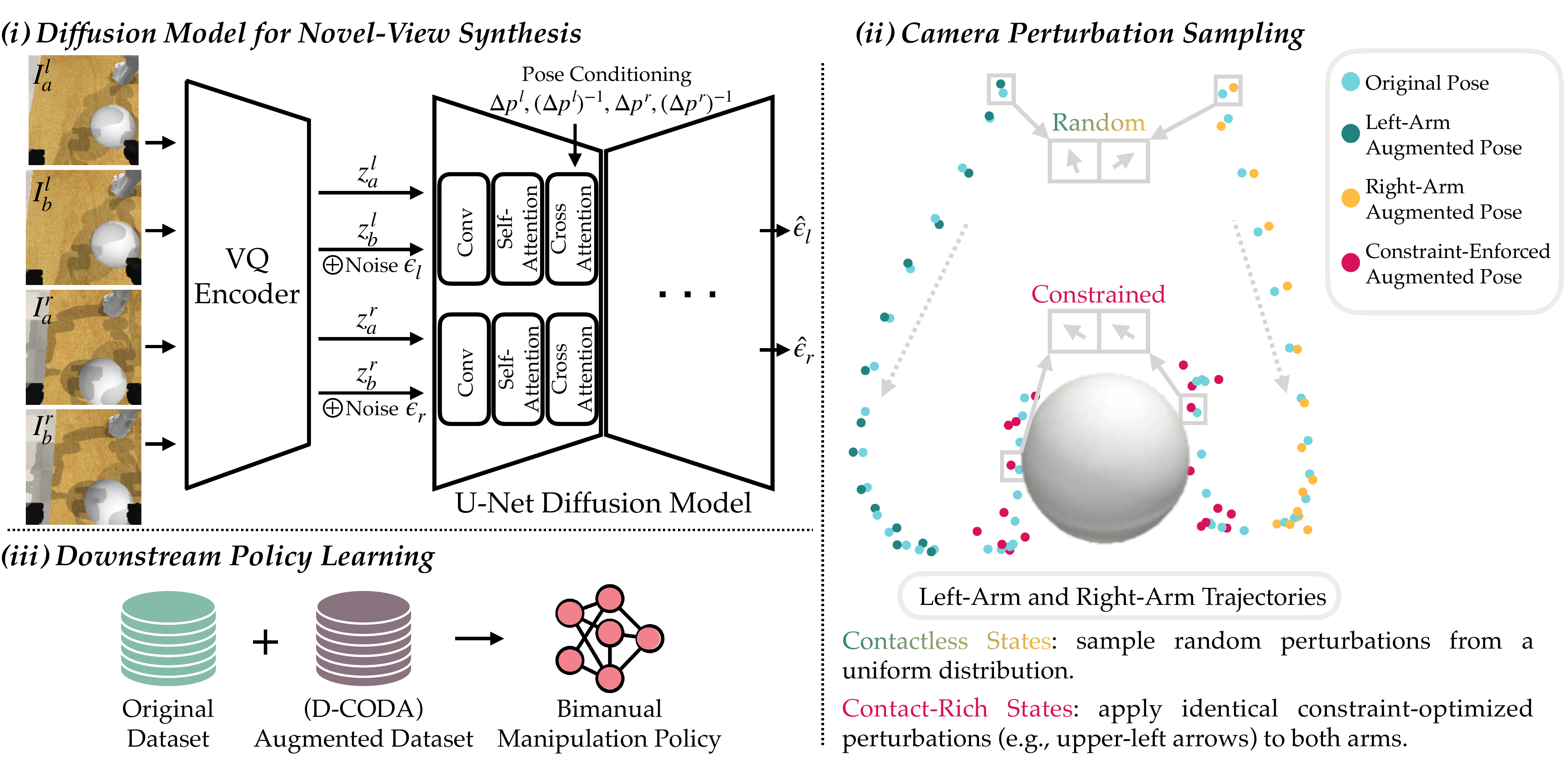} 
\caption{
\textbf{Overview of \method.} \textbf{(i):} The diffusion model is an iterative denoiser that learns to map source wrist-camera images $I_a^l$ and $I_a^r$ to target wrist-camera images $I_b^l$ and $I_b^r$, conditioned on pose transformations $\Delta p^{l}$ and $\Delta p^{r}$, using the original dataset (i.e., the dataset to be augmented). \textbf{(ii):} We use SAM2~\cite{ravi2024sam} to decompose a bimanual manipulation task into contactless and contact-rich states. We uniformly sample random camera pose perturbations for contactless states (\textcolor{figgreen}{green} and \textcolor{figyellow}{yellow} dots). For contact-rich states (\textcolor{figmaroon}{maroon} dots), we use constrained optimization to sample perturbations that satisfy a set of constraints suitable for coordinated manipulation. We then employ the trained diffusion model to synthesize novel views based on the original dataset, using its images and corresponding sampled perturbations. This generates an augmented dataset. \textbf{(iii):} We combine the original and augmented datasets to train a bimanual manipulation policy.}
\label{fig:method}
\vspace*{-10pt}
\end{figure*}

We introduce \method, a diffusion-based framework for data augmentation of eye-in-hand bimanual imitation learning, which synthesizes novel wrist-camera views with action labels (see Figure~\ref{fig:method}). 

\subsection{Diffusion Model for Novel-View Synthesis}
\label{sec:diffusion-model}

We modify the conditional diffusion model proposed by Zhang et al.~\cite{zhang2024diffusionmeetsdagger} to synthesize novel wrist camera views for the two arms. The diffusion model, denoted as $\epsilon_\phi$, is an iterative denoiser. It is conditioned on the source images $I_a^l$ and $I_a^r$ and the pose transformations $\Delta p^{l}$ and $\Delta p^{r}$. The diffusion targets are $I_b^l$ and $I_b^r$. Both source and target images are passed through a VQ-GAN autoencoder $V$~\cite{esser2021taming,rombach2022high} to allow denoising on the latent representations \{$z_a^l$, $z_{b,t}^l$, $z_a^r$, $z_{b,t}^r$\}, which correspond to the source and target images of both robot arms. This enables the diffusion process to operate in the latent space of the autoencoder rather than the high-dimensional pixel space. The model is trained to predict $\hat{\epsilon}_l$ and $\hat{\epsilon}_r$, which correspond to the noise terms $\epsilon_l$ and $\epsilon_r$ that were added to the latent vectors of the noise targets $z_{b,t}^l$ and $z_{b,t}^r$. Thus, the training objective is to minimize $\mathcal{L}$: %

\begin{equation}
\mathcal{L} = \left\| \epsilon_l - \hat{\epsilon}_l \right\|_2^2 
+ \left\| \epsilon_r - \hat{\epsilon}_r \right\|_2^2 \quad \mbox{where} \quad  
\{ \hat{\epsilon}_l, \hat{\epsilon}_r \} = \epsilon_\phi\Big(
z_{b,t}^l, V(I_a^l), \Delta p^{l}, \,
z_{b,t}^r, V(I_a^r), \Delta p^{r}, t \Big)
\end{equation}
and where $z_{b,0}^l = V(I_b^l)$ and $ z_{b,0}^r = V(I_b^r)$.
The diffusion model architecture~\cite{Yu2023PhotoconsistentNVS} is based on U-Net~\cite{ronneberger2015u}, which consists of convolution, cross-attention, and self-attention layers. To condition the model on pose transformations, we inject $\Delta p^{l},(\Delta p^{l})^{-1},\Delta p^{r},(\Delta p^{r})^{-1}$ into the cross-attention layers. This improves the feature representations by incorporating relative camera pose information between the source and target views~\cite{Yu2023PhotoconsistentNVS}. During training, we randomly sample images $\{ I_{a}^l, I_{b}^l, I_{a}^r, I_{b}^r \}$ from a robot trajectory to construct the input $(I_a^l, I_b^l, \Delta p^{l} ,I_a^r,I_b^r,\Delta p^{r})$ for the model, and we compute $\Delta p = {}_aT_b$ by taking the matrix product of the inverse of camera pose $a$ and camera pose $b$.
Given a dataset of expert demonstrations $\mathcal{D}$, we train the diffusion model on $\mathcal{D}$ for a fixed number of iterations. We then use the trained model and sampled camera perturbations (\autoref{sec:camera-perturbation-sampling}) to synthesize novel wrist camera views based on the same dataset.

\subsection{Camera Perturbation Sampling}
\label{sec:camera-perturbation-sampling}

While the prior formulation enables image synthesis, it lacks constraint-enforced action sampling to ensure that sampled perturbations are valid.
We introduce a novel camera pose sampling procedure for coordinated bimanual manipulation tasks.
Given such a task, we decompose it into contactless and contact-rich states.
To detect such contact, we use SAM2~\cite{ravi2024sam} to extract segmentation masks of the grippers and track them throughout the robot trajectories. This approach enables accurate segmentation even when the grippers are closing or partially closed. If depth images from the wrist cameras are available, we use them in conjunction with the masks to detect contact events using z-score filtering, depth thresholding, and mask filling. If depth images are unavailable, we infer contact events by checking whether the grippers are fully visible within the wrist camera view using the Structural Similarity Index (SSIM)~\cite{wang2004image}.

If gripper-to-object contact is absent, we uniformly sample a random direction for each arm within a predefined range of magnitudes [$m_{lb}$, $m_{ub}$] and rotations [$r_{lb}$, $r_{ub}$] to generate camera perturbations. However, when contact is detected, we formulate camera perturbation sampling as a constrained optimization problem. Our key insight is to apply identical perturbations to both arms during contact events, ensuring coordinated behavior. For this, we employ Dual Annealing~\cite{XIANG1997216}, a global optimization algorithm capable of handling constraints, with early stopping. The decision variables are the translation coordinates $c_{\rm trans}$, representing the transformation applied to the camera perturbations (normalized to [-1, 1]). The cost function penalizes perturbations that are too small, and end-effector poses that are either too close to the table or too close to the other end-effector. Additionally, we use an inverse kinematics solver based on the Levenberg-Marquardt (LM) method to check the feasibility of the perturbed end-effector poses. %
We define the overall optimization problem as: 
\[
\begin{aligned}
\underset{c_{\text{trans}}}{\rm minimize} \quad & \text{Cost}(c_{\text{trans}})
\quad \text{subject to} \quad
\begin{cases}
c_{\text{trans}} \in [-1, 1]^3  \quad \mbox{and} \quad c_{\text{trans}} \geq m_{\text{lb}} \\
\text{ProximityToTable}(c_{\text{trans}}) \geq d_{\text{table}} \\
\text{ProximityToOtherEEF}(c_{\text{trans}}) \geq d_{\text{eff}} \\
\text{IKSolver}(c_{\text{trans}}) = \text{valid}
\end{cases}
\end{aligned}
\]
We construct the transformation matrix for the camera perturbation $T$ using the lowest-cost $c_{\rm trans}$ and the identity rotation matrix. In short, this sampling strategy aims to identify a subset of feasible perturbations that better supports coordinated bimanual manipulation tasks.

\subsection{Action Labeling and Dataset Construction}
\label{sec:action-labeling}

\begin{wrapfigure}{R}{0.43\textwidth}
\vspace*{-14pt}
\center
\includegraphics[width=0.42\textwidth]{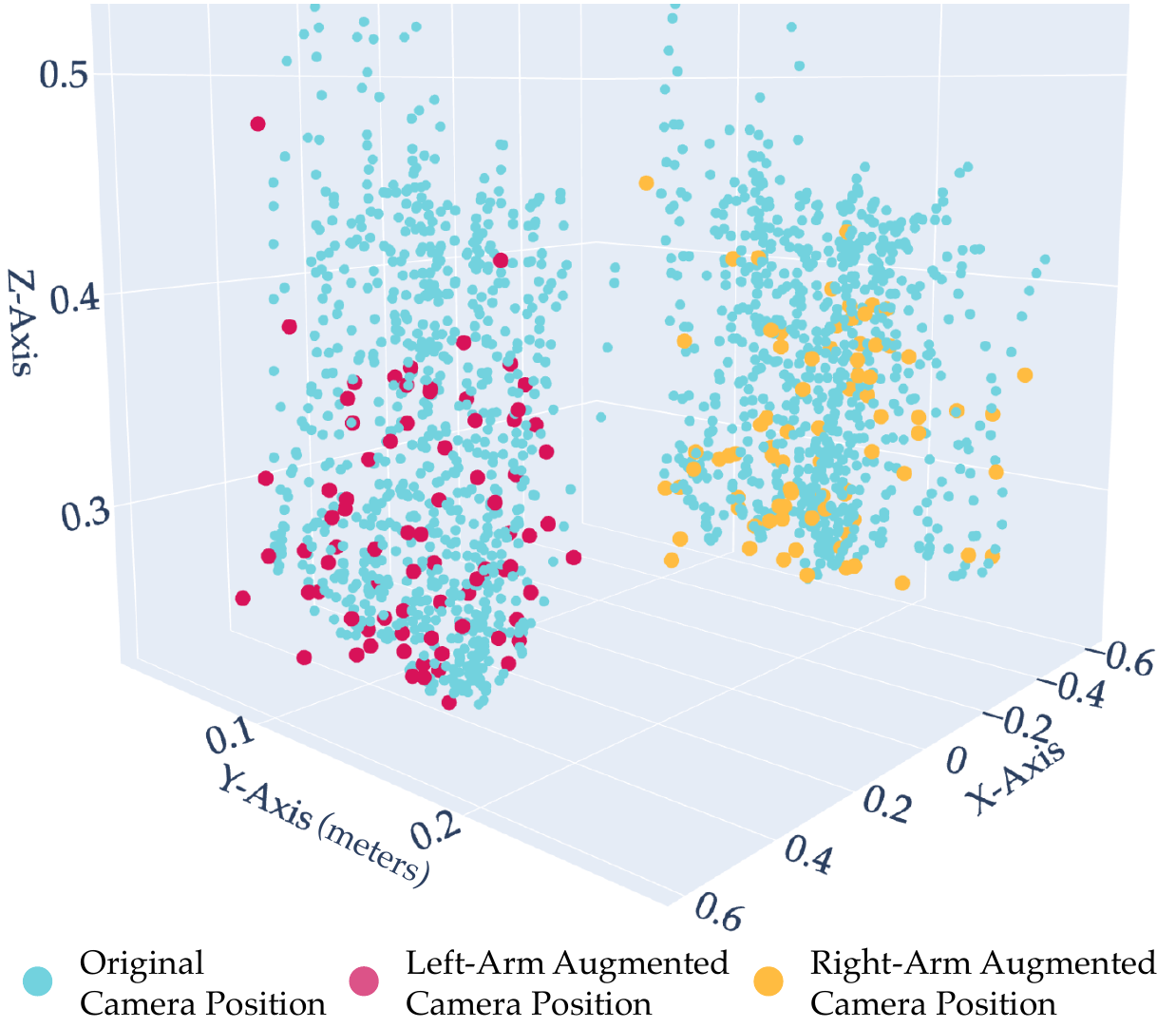} 
\caption{
Isometric view of original and augmented camera positions for the real-world \texttt{Lift Ball} task. The augmented camera positions (\textcolor{figmaroon}{maroon} and \textcolor{figyellow}{yellow} dots) provide broader coverage of state-space regions not occupied by the original camera positions (\textcolor{figblue}{blue} dots).
}
\label{fig:state_coverage}
\vspace*{-12pt}
\end{wrapfigure}
\begin{figure*}[t]
\center
\includegraphics[width=1.0\textwidth]{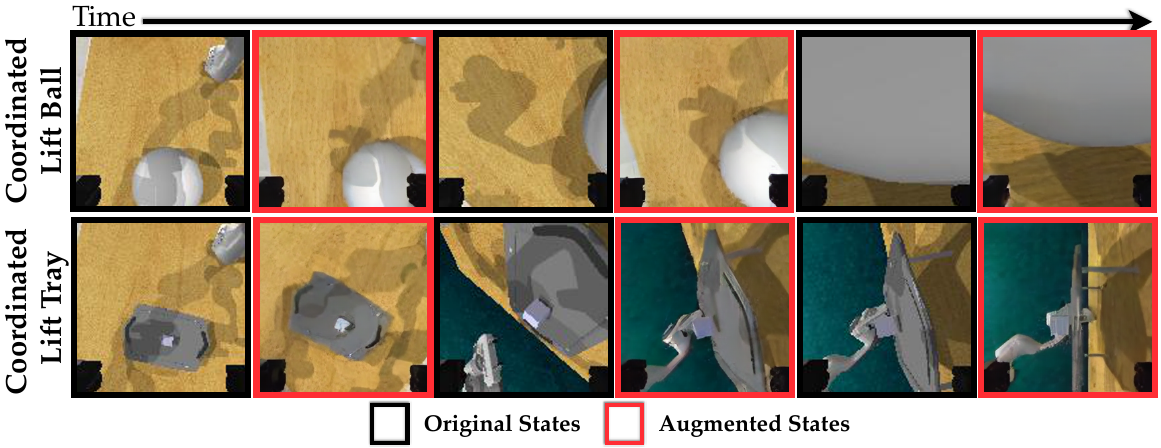} 
\caption{
Examples of the original and synthesized wrist-camera images using \method on \texttt{Coordinated Lift Ball} and \texttt{Coordinated Lift Tray} tasks in simulation. 
The first black column of images are the original states where the following column of \textcolor{figred}{red} images are the augmented (perturbed original) states. All original and augmented state pairs are at the same timestep and each task is from the same episode. See ~\autoref{sec:more-synthesized-images} for more examples.
}
\label{fig:method_qualitative}
\vspace*{-5pt}
\end{figure*}

Given the dataset $\mathcal{D}$ and its corresponding sampled camera perturbations, we use the trained diffusion model to synthesize novel wrist camera views for both arms. To generate perturbed end-effector poses, we perform matrix multiplications involving the camera perturbation transformation $T$, the original camera pose $C$, and the end-effector pose $E$: $C \cdot T \cdot(C)^{-1} \cdot E$.
Since our eye-in-hand imitation learning algorithm operates in joint space, we use the LM inverse kinematics solver to compute the perturbed target joint positions $\tilde{\ba}_t$ ($\tilde{\ba}_t^l$ and $\tilde{\ba}_t^r$). If the resulting configuration is invalid, we discard the augmentation for that state and retain the original state information. Otherwise, we replace the original state with the augmented (out-of-distribution) state every $k$ timesteps, which helps mitigate the issue of compounding errors in behavior cloning policies. The non-augmented action labels and corresponding states remain in-distribution, which guides the behavior cloning policies to complete the tasks.
This will result in an augmented dataset of novel views $\tilde{\mathcal{D}}$, and $\pi_\theta$ is trained on $\mathcal{D} = \mathcal{D} \cup \tilde{\mathcal{D}}$.
See Figure~\ref{fig:state_coverage} for a visualization of the original and augmented camera positions in a combined dataset, and Figure~\ref{fig:method_qualitative} for examples of the synthesized images.
    
\vspace*{-12pt}
\section{Experiments and Results}
\label{sec:result}

\textbf{Simulation.} We adopt five bimanual tasks from PerAct2~\cite{peract2}, which is built on top of RLBench~\cite{james2019rlbench}, a popular robot manipulation benchmark. To improve the performance of the ACT baseline~\cite{Zhao-RSS-23}, we simplify certain tasks by reducing the axes of variation (e.g., shrinking the workspace), as mentioned below.
We use the following five bimanual tasks (see the Appendix for more details): %
\vspace{-4pt}
\begin{itemize}[noitemsep,leftmargin=*]
\item \texttt{\textbf{Coordinated Lift Ball}}: a ball is randomly spawned in a workspace of $0.65 \times 0.91$ \SI{}{\meter}, same as PerAct2. A success is when the ball is lifted to a height above \SI{0.95}{\meter}.
\item \texttt{\textbf{Coordinated Lift Tray Easy}}: a tray with an item placed on top is randomly spawned in a workspace of $0.46 \times 0.64$ \SI{}{\meter}, 70\% of PerAct2's original workspace, and the tray does not rotate. A success is when the tray and the item are lifted to a height above \SI{1.2}{\meter}.
\item \texttt{\textbf{Coordinated Push Box Easy}}: a large box and a target area are randomly spawned in a workspace of $0.59 \times 0.82$ \SI{}{\meter}, 90\% of PerAct2's original workspace, and the box may be randomly rotated by up to 4 degrees. A success is when the box reaches the target area.
\item \texttt{\textbf{Dual Push Buttons}}: three buttons with different colors are randomly spawned in a workspace of $0.65 \times 0.91$ \SI{}{\meter}, as in PerAct2. A success is when two specified buttons are pressed simultaneously. %
\item \texttt{\textbf{Bimanual Straighten Rope}}: a long rope and target areas are randomly spawned in a workspace of $0.65 \times 0.91$ \SI{}{\meter}, same as PerAct2. A success is when both ends of a rope are in their target areas. %
\end{itemize}

\textbf{Real-World.}
We use three coordinated tasks: \texttt{Lift Ball}, \texttt{Lift Drawer}, and \texttt{Push Block}, as coordinated dual-arm data augmentation is our primary focus. 
We use two CB2 UR5 6-DOF robot arms in a bimanual setup in a $0.97 \times 0.79$ \SI{}{\meter} workspace, with a front camera and two wrist-mounted cameras. Each arm has a Robotiq 2F-85 parallel-jaw gripper and an Intel RealSense D415 RGB-D wrist camera.
An experienced roboticist uses GELLO~\cite{wu2024gello} to teleoperate the robots and collects $\sim$32 demonstrations per task. 
For evaluation, we perform 20 rollouts per task. In \texttt{Lift Ball}, a \SI{0.35}{\meter} diameter ball is placed in randomized positions within a $0.64 \times 0.20$ \SI{}{\meter} region. A success is when the ball is lifted to a height above \SI{0.25}{\meter}. In \texttt{Lift Drawer}, a $0.29 \times 0.29 \times 0.15$ \SI{}{\meter} square drawer is placed randomly within a $0.48 \times 0.38$ \SI{}{\meter} region and randomly rotated up to \ang{25}. A success is when the drawer is lifted to a height above \SI{0.22}{\meter}. In \texttt{Push Block}, a $0.07 \times 0.35 \times 0.12$ \SI{}{\meter} foam block is randomly placed within a $0.97 \times 0.43$ \SI{}{\meter} region and rotated up to \ang{13}. A success is when the block is pushed past the front of the workspace.

\subsection{Baselines}
\label{ssec:baselines_ablations}

In simulation, we compare \method against strong baselines: \textbf{Fine-tuned VISTA}~\cite{tian2024vista} and \textbf{Bimanual DMD}. All methods generate an augmented dataset, and we train \textbf{Action Chunking with Transformers (ACT)}~\cite{Zhao-RSS-23}, a state-of-the-art imitation learning method for bimanual manipulation, on both the augmented and original data to evaluate task performance. For all baselines, we adopt PerAct2's ACT implementation with fine-tuned action chunk sizes.
VISTA leverages a diffusion-based novel view synthesis model, ZeroNVS~\cite{sargent2024zeronvs}, to augment third-person viewpoints from a single third-person view. We fine-tune VISTA on each task’s training data using 10 randomly sampled overhead camera viewpoints drawn from a quarter-circle arc distribution. Following the training strategy of the best-performing VISTA variant~\cite{tian2024vista}, we train ACT on both the augmented overhead camera images and the original wrist-camera images.
The Bimanual DMD baseline uses one DMD model per arm to synthesize wrist-camera images and employs the same $k$, interval at which original states are replaced, and random seed as \method to generate perturbed actions and augmented states.
The ACT (w/o augment.) baseline is trained only on the original dataset, serving as a reference for ACT performance without data augmentation.
The ACT (more data) is trained on the original dataset along with 100 additional demonstrations without data augmentation, serving as an upper bound for ACT performance with more expert data.

\subsection{Experiment Protocol and Evaluation}

In simulation, we use the same training, validation, and testing data with the same environment seeds across all methods to ensure a fair comparison. Demonstrations are generated using a waypoint-based motion planner in RLBench~\cite{james2019rlbench}. We train the ACT policy for all methods using 100 episodes of training data along with their corresponding augmented data, saving a checkpoint every 2,000 iterations up to a total of 260,000 iterations. All checkpoints are validated using the same 25 episodes of validation data. Based on validation performance, the best-performing checkpoint is then evaluated on 25 unseen test data. In real-world experiments, we use the last checkpoint for each method and attempt to use the same starting configurations (e.g., object spawn locations and rotations).

\subsection{Simulation Results}

\begin{table*}[t]
  \setlength\tabcolsep{4.6pt}
  \centering
    \footnotesize
    \begin{tabular}{lccccccccccc}
    \toprule
        & \multicolumn{1}{c}{\# of} & \multicolumn{1}{c}{Coordinated} & \multicolumn{1}{c}{Coordinated} & \multicolumn{1}{c}{Coordinated} & \multicolumn{1}{c}{Dual Push} & \multicolumn{1}{c}{Bimanual} \\ 
        Method & \multicolumn{1}{c}{Cameras} & \multicolumn{1}{c}{Lift Ball} & \multicolumn{1}{c}{Lift Tray} & \multicolumn{1}{c}{Push Box} & \multicolumn{1}{c}{Buttons} & \multicolumn{1}{c}{Straighten Rope} \\ 
    \midrule
    Fine-tuned VISTA & 3 & 61.3 & 2.6 & \textbf{76.0} & 1.3 & 26.7 \\
    \rowcolor{lightgreen} \method (ours) & 3 & \textbf{77.3} & 34.7 & 58.7 & 34.7 & \textbf{48.0}   \\ 
    \midrule
    Bimanual DMD & 2 & 50.7 & 13.3 & 32.0 & 48.0 & 13.0 \\
    ACT (more data) & 2 & 48.0 & 26.7 & 29.3 & 49.3 & 26.7 \\
    ACT (w/o augment.) & 2 & 56.0 & 37.3 & 36.0 & 46.7 & 18.7 \\ 
    \rowcolor{lightgreen} \method (ours) & 2 & 73.3 & \textbf{44.0} & 56.0 & \textbf{53.3} & 30.7   \\ 
    \bottomrule
    \end{tabular}
  \caption{
    Results from simulation experiments comparing \method against four baselines (see Section~\ref{ssec:baselines_ablations}). The success rate results are the average evaluation over three seeds. The ACT policy is used across all methods.
  }
  \vspace*{-10pt}
  \label{tab:sim-results}
\end{table*}

Table~\ref{tab:sim-results} reports the test success rates of different methods in simulation. \method outperforms the baselines on 4 out of 5 tasks, including non-coordinated tasks such as \texttt{Dual Push Buttons} and \texttt{Bimanual Straighten Rope}. However, its performance is lower than VISTA on \texttt{Coordinated Push Box} because wrist-camera views offer poor visibility of the scene (i.e., the position of the box relative to the target area). As a result, augmenting wrist-camera views does not significantly improve the ACT baseline performance, although we still achieve a 20\% improvement.
Qualitatively, all methods can fail due to imprecise grasping, pushing, or placing of objects. Baseline methods, particularly VISTA, struggle with tasks that require a low tolerance for error, such as grasping tray handles or pushing small buttons. Overall, \method makes fewer errors in these scenarios. We also observe that both Bimanual DMD and \method, which generate out-of-distribution states, learn to recover from failures. For instance, when the grippers slide off the box during the \texttt{Coordinated Push Box} task, they recover by repositioning the grippers and continuing to push the box to complete the task.
Another interesting observation is that all methods using an overhead camera have worse performance by the downstream ACT policy on high-precision tasks (e.g., \texttt{Coordinated Lift Tray} and \texttt{Dual Push Buttons}) compared to ACT without using the overhead camera. Therefore, we suspect this limitation is from the design of the downstream ACT policy rather than the data augmentation methods.
Further, we found that learning from more data, ACT (more data), does not always improve policy performance, and data augmentation shows potential for improving performance.
See the Appendix for details of our ablation study.

\subsection{Real-World Results}
\label{ssec:real-world-results}
\vspace*{-1pt}

\begin{wraptable}{r}{0.58\textwidth}
  \vspace*{-30pt}
  \setlength\tabcolsep{3.0pt}
  \centering
    \footnotesize
    \begin{tabular}{lcccccccccc}
    \toprule
        Method & \multicolumn{1}{c}{Lift Ball} & \multicolumn{1}{c}{Lift Drawer} & \multicolumn{1}{c}{Push Block} \\ 
    \midrule
    Fine-tuned VISTA & 12 / 20 & 0 / 20 & \textbf{20 / 20} \\
    ACT (w/o augment.) & 15 / 20 & 7 / 20 & 15 / 20 \\ 
    \rowcolor{lightgreen} \method (ours) & \textbf{17 / 20} & \textbf{14 / 20} & \textbf{20 / 20} \\ 
    \midrule
    $\pi_0$-FAST (w/o augment.) & 2 / 20 & \textbf{1 / 20} & \textbf{20 / 20} \\ 
    \rowcolor{lightgreen} \method (ours) & \textbf{12 / 20} & \textbf{1 / 20} & \textbf{20 / 20} \\ 
    \bottomrule
    \end{tabular}
  \caption{
    Real-world experiment results comparing \method with baselines, with 20 trials per method and task combination. 
  }
  \label{tab:real-results}
  \vspace{-10pt}
\end{wraptable}

Real-world results are shown in Table~\ref{tab:real-results} and example rollouts in Figure~\ref{fig:real_tasks_and_rollouts}. The top three rows use ACT as the downstream manipulation policy, while the bottom two use $\pi_0$-FAST~\cite{pertsch2025fast}, a vision-language-action model.
$\pi_0$-FAST is fine-tuned using Low-Rank Adaptation (LoRA) with the Gemma-2B-LoRA variant for 150,000 training steps, provided in~\cite{pertsch2025fast}. 
Fine-tuned VISTA uses all three cameras, following its best-performing variant, whereas the other methods use only the wrist cameras, except in \texttt{Push Block}, where we found that all methods benefit from third-person views.
\method outperforms baselines on all three tasks based on evaluations over 20 trials.

We observe that in \texttt{Lift Ball}, when using ACT, the robot arms freeze less frequently with \method compared to baselines when the arms are in contact and lifting the ball.
When using $\pi_0$-FAST, the baseline frequently misses the ball by moving the arms over it, or squeezes the ball so tightly that it triggers a force limit error on the robot.
In contrast, \method more reliably completes the task by positioning the arms beneath the ball to lift it, a strategy not seen in the baseline.
However, most failures of \method are due to large action values generated from the policy, causing the arms to deviate from the intended trajectory. We suspect that the large actions may result from the discontinuous nature of action tokens, as the augmented states are out-of-distribution, perturbed original states, which could inadvertently cause the policy to learn to output actions that suddenly deviate from the trajectory. This issue does not appear in ACT and might be mitigated by adopting a smoother action token representation.
In \texttt{Push Block}, the robot arms using \method get stuck less often when the block is positioned farther from the grippers, compared to the ACT baseline.
In \texttt{Lift Drawer}, our method reaches the sides of the drawer more frequently than the baselines, an intermediate subgoal necessary to complete the task.
Fine-tuned VISTA performs very poorly on this task, similar to its performance in \texttt{Coordinated Lift Tray Easy}.
In 9 out of 20 trials, VISTA successfully reached the sides of the drawer, but the policy failed to close the grippers.
These results suggest that VISTA struggles with tasks requiring precise manipulation, as shown in both simulation and real-world experiments. We suspect that this limitation arises from the use of augmentations for third-person views, which may adversely affect policy learning when wrist-camera views are more critical for task success. In other words, the policy may prioritize learning invariant features from third-person views rather than focusing on task-relevant features in the wrist-camera views.
Overall, both $\pi_0$-FAST and ACT demonstrate improved performance with our data augmentation; however, with limited training data, ACT appears to exhibit greater reliability.

\section{Conclusion}
\label{sec:conclusion}

In this paper, we study data augmentation for bimanual manipulation, focusing specifically on eye-in-hand bimanual imitation learning. Our method, \method, uses a diffusion model to generate diverse and consistent wrist camera images while enforcing and generating appropriate action labels using constrained optimization. By augmenting data, we obtain improved imitation learning performance across a range of diverse bimanual tasks. We hope our work inspires future exploration of data augmentation methods for bimanual manipulation.

\section{Limitations}
\label{sec:limitations}
While promising, \method has limitations that suggest opportunities for future work. First, our method is limited to augmenting wrist view images and is not intended for third-person view augmentation. Augmenting third-person views while modifying the action labels is nontrivial, as it requires the augmented views to reflect the change in movements of the robot arms implied by the augmented action labels. Another limitation is that our method relies on the distribution of novel camera poses being ``sufficiently similar'' to those in training, and would likely suffer with substantially different camera poses. Finally, although using \method improves downstream policy performance by reducing the number of failures, it does not completely eliminate them.

\section{Acknowledgments}
\label{sec:acknowledgments}
We thank our colleagues from the Robotic Embedded Systems Laboratory (RESL) and the Sensing, Learning, and Understanding for Robotic Manipulation (SLURM) lab at the University of Southern California for their fruitful discussions and helpful writing feedback.

\bibliography{example}  %

\clearpage 
\appendix

\section{Paper Changelog}
Version 1 on arXiv was the initial public release of the paper. Version 2 included minor writing and BibTeX edits and served as the camera-ready version for CoRL 2025.

\section{Task Details}

\begin{figure*}[h]
\center
\includegraphics[width=1.0\textwidth]{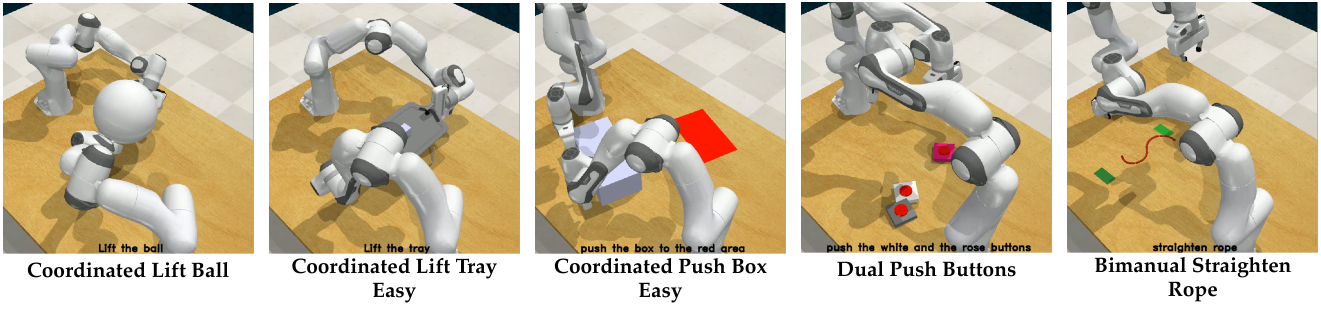} 
\caption{
Simulation environments for our bimanual manipulation tasks, adapted from PerAct2~\cite{peract2}.
}
\label{fig:tasks}
\vspace*{-10pt}
\end{figure*}  

\begin{figure*}[h]
\centering

\includegraphics[width=1.0\textwidth]{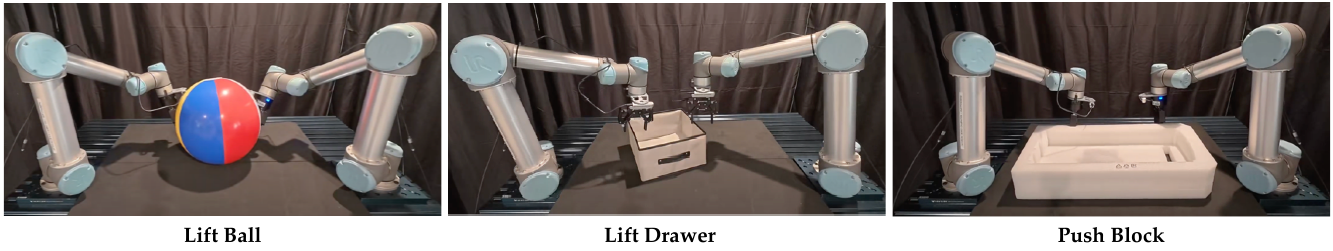} 
\vspace{4pt} %

\includegraphics[width=1.0\textwidth]{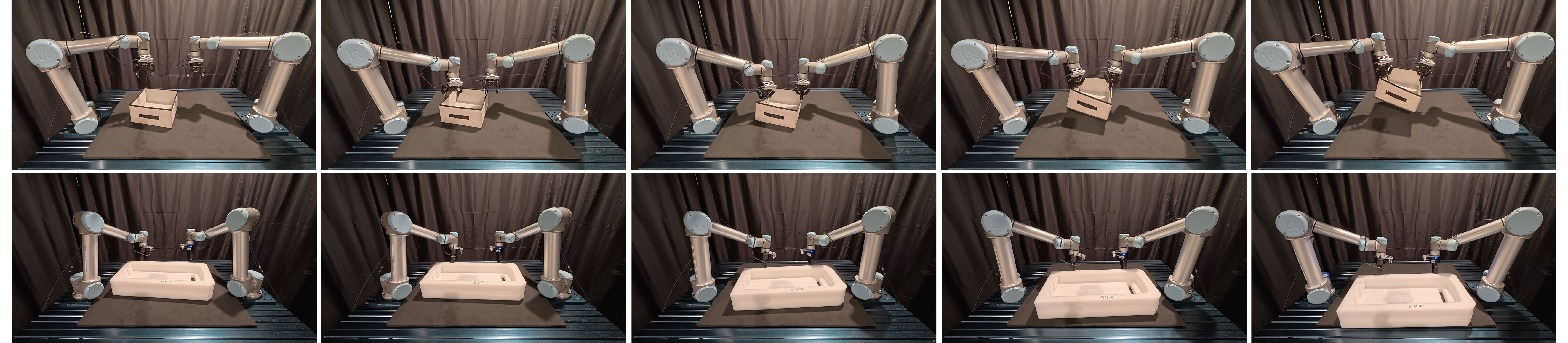}

\caption{
Top: Real-world bimanual manipulation tasks.  
Bottom: Example successful rollouts (\texttt{Lift Drawer} on top row; \texttt{Push Block} on bottom row) of \method on a real-world bimanual setup with UR5s. See Section~\ref{ssec:real-world-results} for quantitative results.
}
\label{fig:real_tasks_and_rollouts}
\vspace*{-10pt}
\end{figure*}

\begin{figure*}[h]
\center
\includegraphics[width=1.0\textwidth]{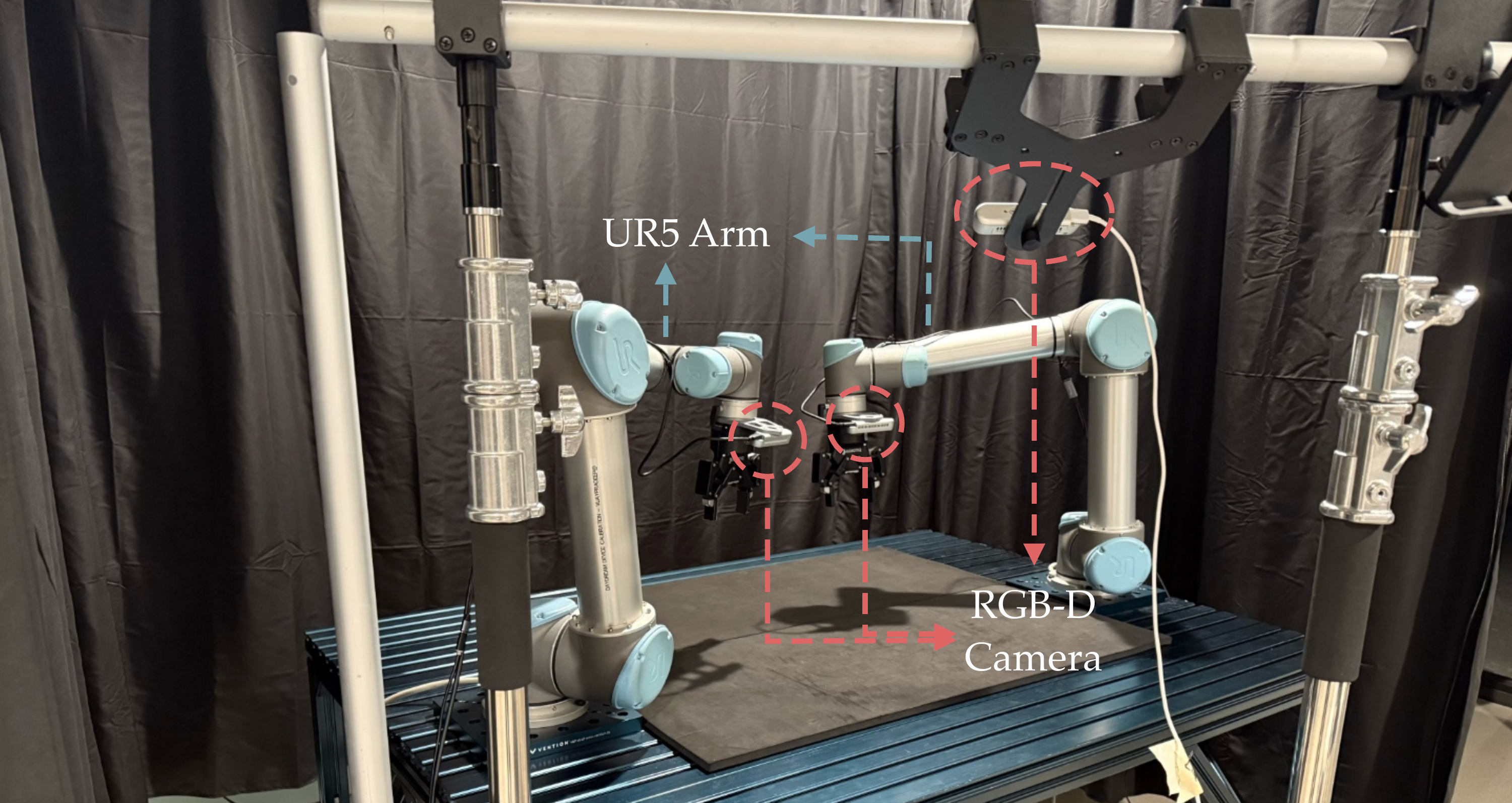} 
\caption{
Real-world bimanual UR5 setup.
}
\label{fig:tasks}
\vspace*{-10pt}
\end{figure*}

\newpage

\section{Ablations}

\textbf{Ablations of \method.} In simulation, we test the following methods:
\vspace{-4pt}
\begin{itemize}[noitemsep,leftmargin=*]
\item \textbf{\method with Replaced Encoders}: uses a VQGAN encoder trained on the Open Images~\cite{kuznetsova2018open} dataset from Latent Diffusion~\cite{rombach2022high} instead of the RealEstate10K~\cite{zhou2018stereo} dataset.
\item \textbf{\method w/o Constrained Optim.}: does not use constrained optimization to sample camera perturbations (i.e., random sampling).
\end{itemize}

\begin{wraptable}{r}{0.47\textwidth}
  \setlength\tabcolsep{4.0pt}
  \centering
    \footnotesize
    \begin{tabular}{lccccccc}
    \toprule
        & \multicolumn{1}{c}{Coordinated} \\
    Method & \multicolumn{1}{c}{Lift Ball} \\ 
    \midrule
    \method with Replaced Encoders & 53.3 \\
    \method w/o Constrained Optim. & 57.3 \\ 
    \rowcolor{lightgreen} \method (ours) & \textbf{73.3} \\ 
    \bottomrule
    \end{tabular}
  \caption{
  Ablation experiment results in simulation.
  }
  \vspace*{-10pt}
  \label{tab:abl-results}
\end{wraptable}
Table~\ref{tab:abl-results} indicates that \method performs best with constrained optimization and the original VQGAN encoder. \method without constrained optimization fails more often than \method during ball lifting, as expected, since the perturbations generated by the model are not constraint-enforced and are entirely random. See Section~\ref{sec:constraint-enforced-action-sampling} for details on the importance of constraint-enforced action sampling. Additionally, the model with replaced encoders generates images with more artifacts, resulting in poorer policy performance.

\section{Generalization Experiment}
\begin{wraptable}{r}{0.47\textwidth}
  \vspace*{-9pt}
  \setlength\tabcolsep{4.0pt}
  \centering
    \footnotesize
    \begin{tabular}{lccccccc}
    \toprule
        & \multicolumn{1}{c}{Coordinated} \\
    Method & \multicolumn{1}{c}{Lift Ball} \\ 
    \midrule
    Zero-Shot & 44.0 \\
    Few-Shot (10 demos) & 60.0 \\ 
    Train from Scratch (100 demos) & \textbf{73.3} \\ 
    \bottomrule
    \end{tabular}
  \caption{
  Generalization experiment results in simulation.
  }
  \vspace*{-5pt}
  \label{tab:generalization}
\end{wraptable}
We evaluate the diffusion model’s generalization capability to unseen objects and tasks. For the zero-shot and few-shot experiments, we train the model on 100 demonstrations each from the following PerAct2~\cite{peract2} tasks: \texttt{Coordinated Lift Tray}, \texttt{Pick Up Notebook}, \texttt{Pick Up Plate}, \texttt{Sweep Dust Pan}, and \texttt{Coordinated Push Box}. We then use the trained model to synthesize images for the \texttt{Coordinated Lift Ball} dataset. The following methods are tested:

\begin{itemize}[noitemsep,leftmargin=*]
\item \textbf{Zero-Shot}: uses the trained diffusion model to synthesize images without any fine-tuning.
\item \textbf{Few-Shot (10 demos)}: fine-tunes the trained model for 3000 additional epochs using 10 demonstrations from the target \texttt{Coordinated Lift Ball} dataset, which is then use for image synthesis.
\item \textbf{Train from Scratch (100 demos)}: Trains the diffusion model directly on 100 demonstrations from the \texttt{Coordinated Lift Ball} dataset, without using demonstrations from other tasks.
\end{itemize}

As shown in Table~\ref{tab:generalization}, the diffusion model performs best when trained directly on the target dataset (i.e., the dataset to be augmented). However, when data collection for the target task is costly, the model still achieves reasonable performance in the few-shot setting. Qualitatively, the images synthesized by the model trained from scratch contain the fewest artifacts, with image quality degrading as fewer target demonstrations are used during training.

\section{Additional Implementation Details}

For training the diffusion model, we use the same VQ-GAN pre-trained checkpoint at 2000 epochs with frozen codebooks as DMD~\cite{zhang2024diffusionmeetsdagger}. To randomly sample images ${ I_{a}^l, I_{b}^l, I_{a}^r, I_{b}^r }$ from a robot trajectory to construct the input $(I_a^l, I_b^l, \Delta p^{l}, I_a^r, I_b^r, \Delta p^{r})$, we sample from a range of $\{5,\ldots,15\}$ for all simulation tasks, except for \texttt{Dual Push Buttons}, where we use $\{10,\ldots,30\}$. For real-world experiments, we use a range of $\{1,\ldots,3\}$ for all tasks. For example, if ${I_{a}^l, I_{a}^r}$ are from timestep $t$, then ${I_{b}^l, I_{b}^r}$ are sampled from a future timestep between $t+1$ and $t+3$ in real-world tasks. We use two NVIDIA 4060 Ti GPUs for both training the diffusion model and performing image synthesis.

\begin{wrapfigure}{r}{0.38\textwidth}
  \centering
  \includegraphics[width=0.38\textwidth]{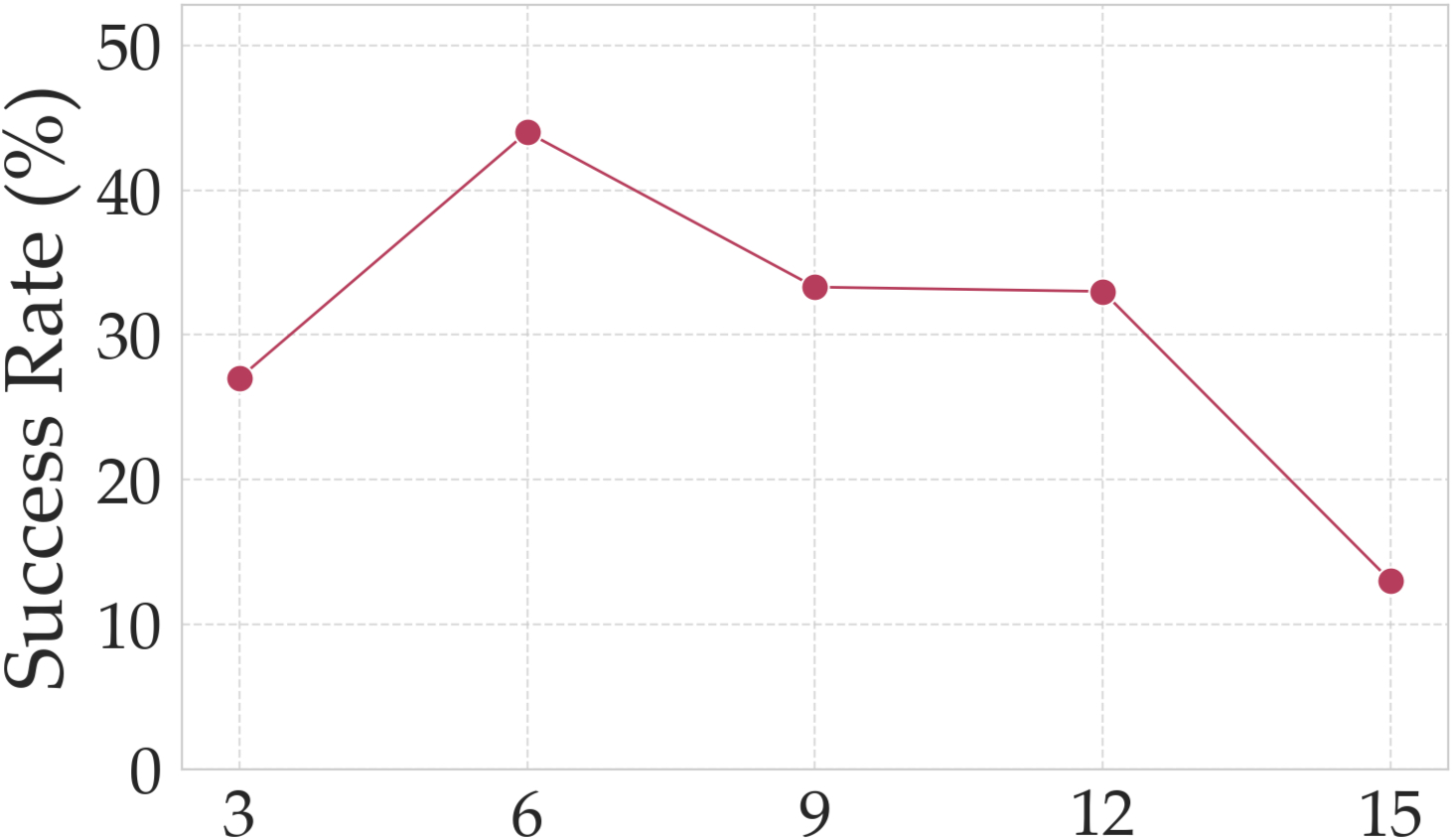}
  \caption{Effects of $k$ on downstream ACT performance on \texttt{Coordinated Lift Tray Easy}.}
  \label{fig:plot-k}
\end{wrapfigure}

For camera perturbation sampling, the translation magnitudes [$m_{lb}$, $m_{ub}$] are set to 0.01 and 0.02 meters, respectively, for contactless and contact-rich states. For contactless states, the rotation bounds [$r_{lb}$, $r_{ub}$] are set to –28.7 and 28.7 degrees, respectively.
For $k$ (i.e., the interval at which original states are replaced), we set $k=6$ for \texttt{Coordinated Lift Ball}, \texttt{Coordinated Lift Tray Easy}, \texttt{Dual Push Buttons}, and all real-world tasks, and $k=9$ for \texttt{Coordinated Push Box Easy} and \texttt{Bimanual Straighten Rope}. Figure~\ref{fig:plot-k} shows that our method is largely insensitive to the choice of $k$, which motivates our choice of 6 as the default value for most tasks.

Table~\ref{tab:hparam-act} summarizes the ACT hyperparameters. While PerAct2 uses a default action chunk size of 10, we found it to yield suboptimal performance across most tasks. To address this, we tune the chunk size for all tasks except \texttt{Coordinated Lift Ball}, using a chunk size of 15 for \texttt{Coordinated Lift Tray Easy} and \texttt{Coordinated Push Box Easy}, and 60 for \texttt{Dual Push Buttons} and \texttt{Bimanual Straighten Rope}. We use a chunk size of 2 across all real-world tasks. In both simulation and real-world experiments, the RGB images have dimensions of $128 \times 128$. An NVIDIA 2080 Ti GPU is used to train the ACT policy.

\begin{table*}[h]
  \vspace{-2pt}
  \setlength\tabcolsep{5.0pt}
  \centering
    \footnotesize
    \begin{tabular}{lc}
    \toprule
    Hyperparameter & Value \\ 
    \midrule
    learning rate & 1e-5 \\
    batch size & 16 \\
    \# encoder layers & 4 \\
    \# decoder layers & 7 \\
    feedforward dimension & 3200 \\
    hidden dimension & 512 \\
    \# heads & 8 \\
    beta & 100 \\
    dropout & 0.1 \\
    \bottomrule
    \end{tabular}
  \caption{
    Hyperparameters of ACT
  }
  \label{tab:hparam-act}
  \vspace{-2pt}
\end{table*}

For real-world experiments, we use Intel RealSense D415 cameras to capture RGB images at a resolution of $640 \times 480$ pixels. These images are first zero-padded and then rescaled to $128 \times 128$. We use the \href{https://github.com/SintefManufacturing/python-urx}{python-urx} library to control the robot arms and I/O programming to operate the Robotiq 2F-85 grippers.

\section{Additional Implementation Details for the Baselines}

For fine-tuned VISTA, 10 overhead camera viewpoints are randomly sampled from a quarter-circle arc distribution and are used to train ZeroNVS with VISTA’s default fine-tuning parameters. The ZeroNVS model is fine-tuned for 5,000 steps on four NVIDIA A40 GPUs. The resulting model is then used to synthesize overhead camera views for all timesteps in each episode. These synthesized images replace all the original overhead images and are used to train ACT.

\section{Lack of Constraint-Enforced Action Sampling}
\label{sec:constraint-enforced-action-sampling}
\begin{figure*}[h]
\center
\includegraphics[width=1.0\textwidth]{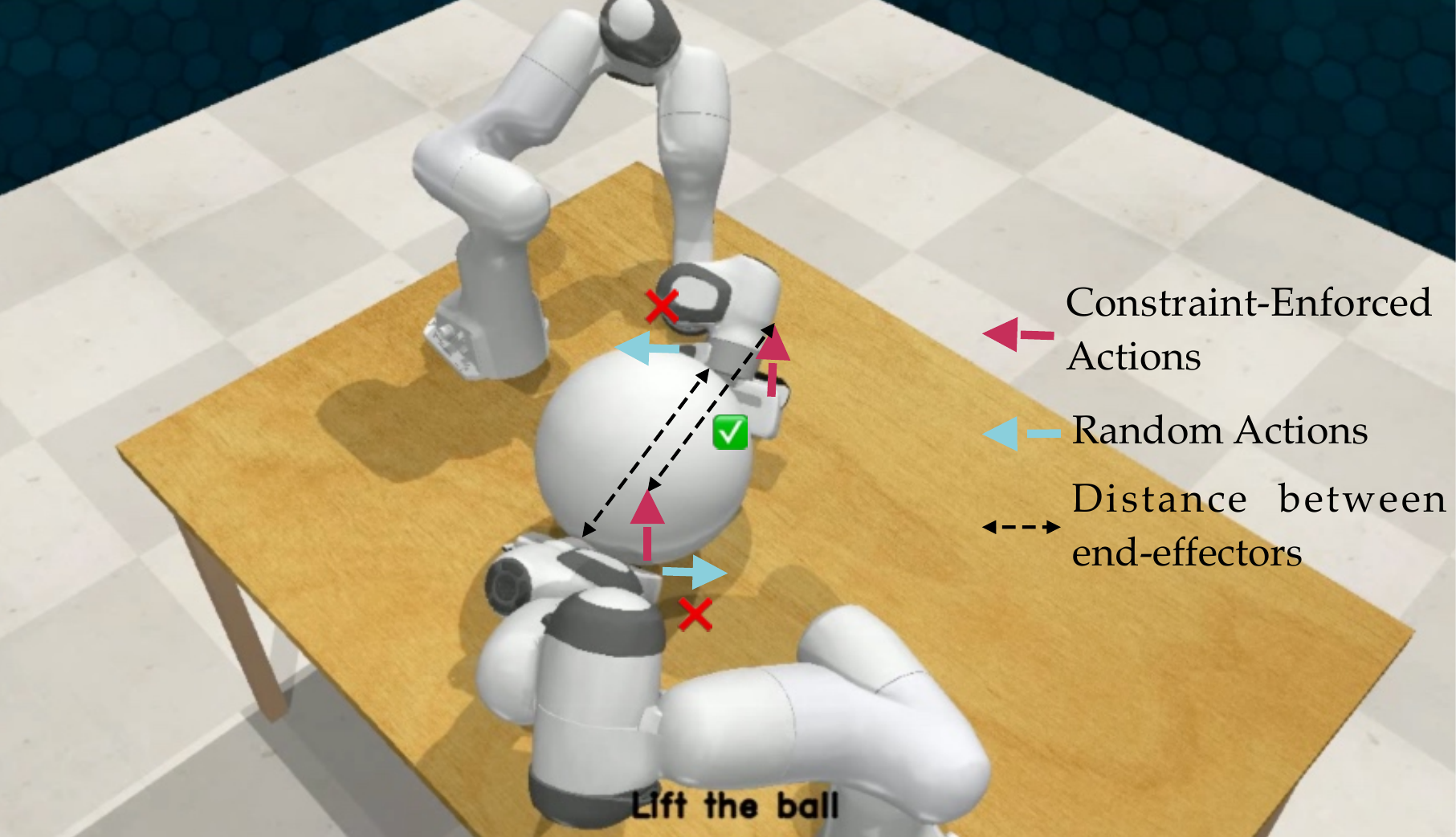} 
\caption{
Visualization comparing constraint-enforced actions and random actions.
}
\label{fig:lack-of-constraint-enforced-actions}
\vspace*{0pt}
\end{figure*}  
Figure~\ref{fig:lack-of-constraint-enforced-actions} shows a visualization comparing constraint-enforced actions and random actions in the \texttt{Coordinated Lift Ball} task. At this timestep, the robot arms have reached the bottom of the ball and are about to lift it. If the next sampled actions are random (\textcolor{figblue}{blue} arrows), they may cause the ball to fall due to an increased distance between the end-effectors (black arrows). In contrast, if constraint-enforced actions are sampled (\textcolor{figmaroon}{maroon} arrows), the distance and orientations of the end-effectors are maintained, preserving the conditions necessary for the ball to remain stable atop the grippers. Thus, constraint-enforced actions are critical for achieving coordinated bimanual manipulation.

\newpage
\section{Examples of Synthesized Images}
\label{sec:more-synthesized-images}
\begin{figure*}[h]
\center
\includegraphics[width=1.0\textwidth]{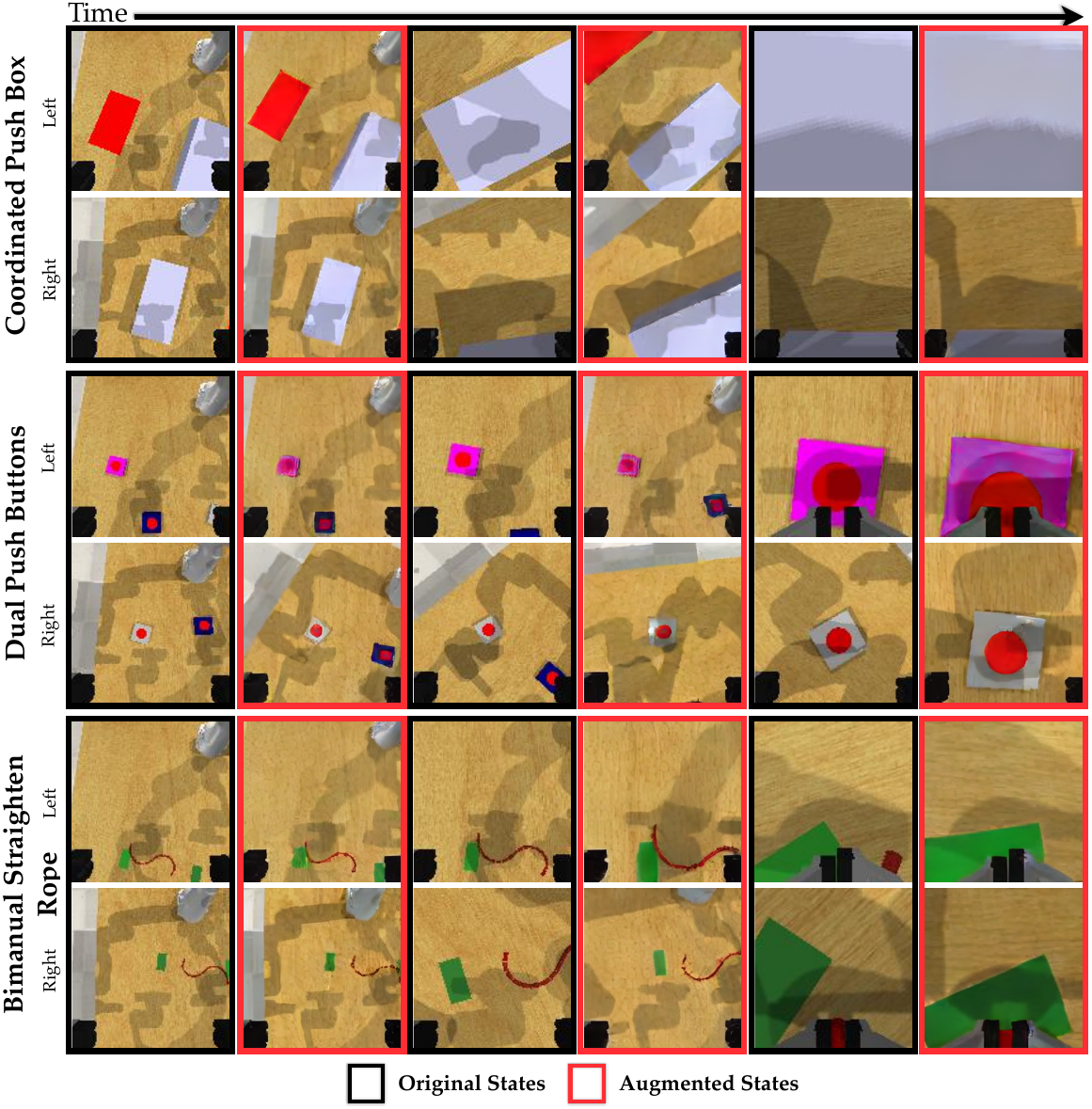} 
\caption{
Examples of the original and synthesized wrist-camera images from both arms using \method on \texttt{Coordinated Push Box}, \texttt{Dual Push Buttons}, and \texttt{Bimanual Straighten Rope} tasks in simulation. 
The first black column of images are the original states where the following column of \textcolor{figred}{red} images are the augmented (perturbed original) states. All original and augmented state pairs are at the same timestep and each task is from the same episode.
}
\label{fig:appendix-simulation-synthesized-images}
\vspace*{0pt}
\end{figure*}

\clearpage
\begin{figure*}[!t]
\centering
\includegraphics[width=1.0\textwidth]{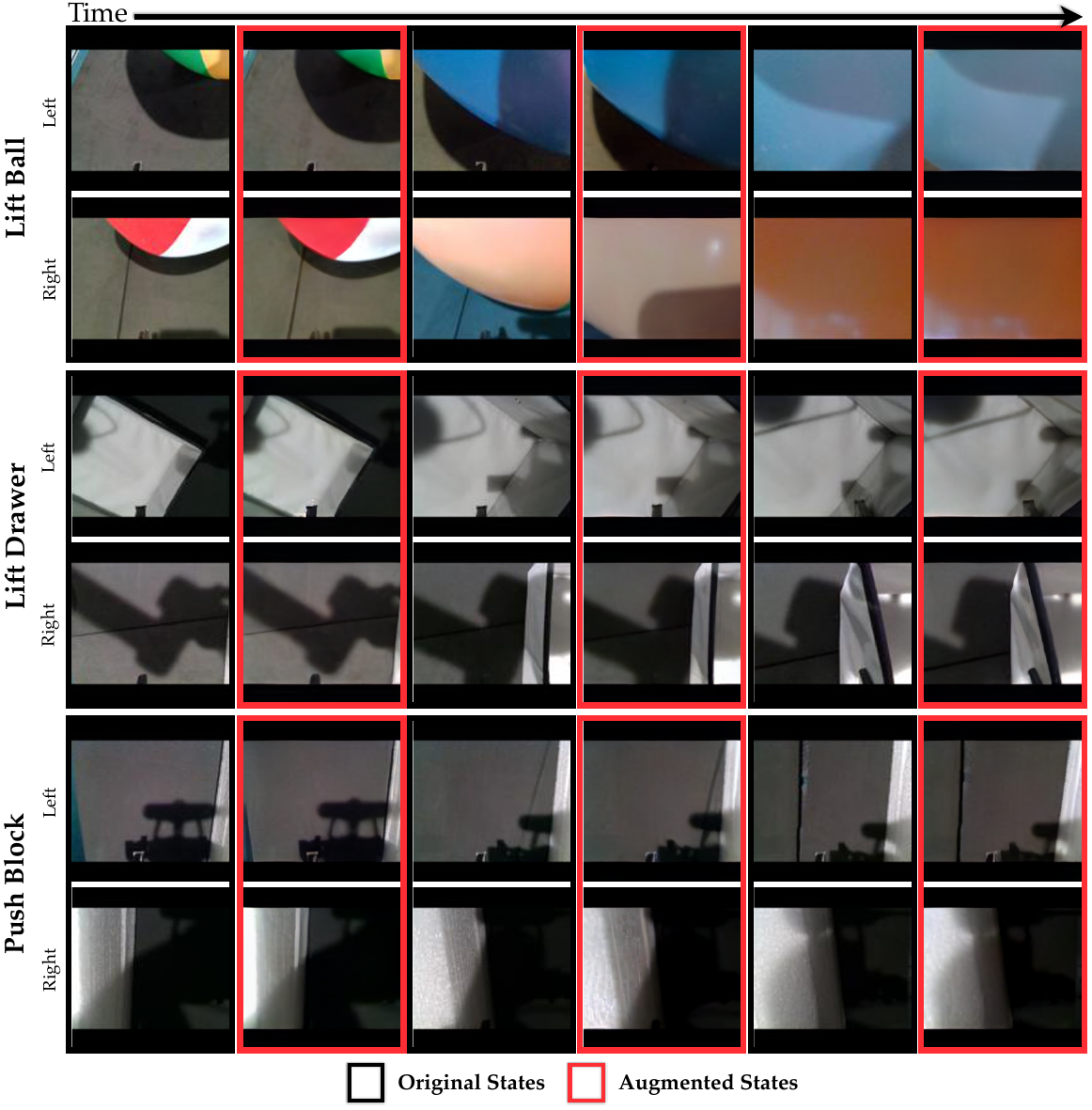} 
\caption{
Examples of the original and synthesized wrist-camera images from both arms using \method on the real-world \texttt{Lift Ball}, \texttt{Lift Drawer}, and \texttt{Push Block} tasks. 
The first black column of images are the original states where the following column of \textcolor{figred}{red} images are the augmented (perturbed original) states. All original and augmented state pairs are at the same timestep and each task is from the same episode.
}
\label{fig:appendix-real-world-synthesized-images}
\vspace*{-10pt}
\end{figure*}

\end{document}